\documentclass[sigconf]{acmart}
\AtBeginDocument{%
  }

\usepackage{amsmath}
\usepackage{bm}
\usepackage{indentfirst}
\usepackage{epsfig}
\usepackage{epstopdf}
\usepackage{url}
\usepackage{xspace}
\usepackage{subeqnarray}
\usepackage{subcaption}
\usepackage{color}
\usepackage{makecell}
\usepackage{stfloats}
\usepackage{multirow}

\newcommand{\paratitle}[1]{\vspace{1.5ex}\noindent\textbf{#1}}
\newcommand{\ie}{\emph{i.e.,}\xspace}

\newcommand{\eg}{\emph{e.g.,}\xspace}

\newcommand{\wrt}{w.r.t.\xspace}
\newcommand{\ignore}[1]{}

\copyrightyear{2025}
\acmYear{2025}
\setcopyright{acmlicensed}
\acmConference[CIKM '25]{Proceedings of the 34th ACM International Conference on Information and Knowledge Management}{November 10--14, 2025}{Seoul, Republic of Korea}
\acmBooktitle{Proceedings of the 34th ACM International Conference on Information and Knowledge Management (CIKM '25), November 10--14, 2025, Seoul, Republic of Korea}
\acmDOI{10.1145/3746252.3760995}
\acmISBN{979-8-4007-2040-6/2025/11}

\begin{document}

\title{STARec: An Efficient Agent Framework for Recommender Systems via Autonomous Deliberate Reasoning}

\author{Chenghao Wu}
\authornote{These authors contributed equally to this research.}
\orcid{0009-0007-9090-6111}
\affiliation{%
  \institution{Gaoling School of Artificial Intelligence, Renmin University of China}
  \city{Beijing}
  \country{China}
}
\email{wuchenghao@ruc.edu.cn}

\author{Ruiyang Ren}
\authornotemark[1]
\orcid{0000-0002-0562-9911}
\affiliation{%
  \institution{Gaoling School of Artificial Intelligence, Renmin University of China}
  \city{Beijing}
  \country{China}
}
\email{reyon.ren@ruc.edu.cn}

\author{Junjie Zhang}
\authornotemark[1]
\orcid{0009-0008-8864-915X}
\affiliation{%
  \institution{Gaoling School of Artificial Intelligence, Renmin University of China}
  \city{Beijing}
  \country{China}
}
\email{junjie.zhang@ruc.edu.cn}

\author{Ruirui Wang}
\orcid{0009-0004-5415-8396}
\affiliation{%
 \institution{Poisson Lab, Huawei}
 \city{Beijing}
 \country{China}}
\email{wangruirui12@huawei.com}

\author{Zhongrui Ma}
\orcid{0009-0002-5907-1607}
\affiliation{%
 \institution{Poisson Lab, Huawei}
 \city{Beijing}
 \country{China}}
\email{zhongrui.ma@huawei.com}

\author{Qi Ye}
\orcid{0009-0006-5907-5746}
\affiliation{%
 \institution{Poisson Lab, Huawei}
 \city{Shenzhen}
 \country{China}}
\email{ye.qi@huawei.com}

\author{Wayne Xin Zhao}
\authornote{Corresponding author.}
\orcid{0000-0002-8333-6196}
\affiliation{%
  \institution{Gaoling School of Artificial Intelligence, Renmin University of China}
  \city{Beijing}
  \country{China}
}
\email{batmanfly@gmail.com}

\renewcommand{\shortauthors}{Chenghao Wu et al.}

\begin{abstract}
While modern recommender systems are instrumental in navigating information abundance, they remain fundamentally limited by static user modeling and reactive decision-making paradigms. Current large language model~(LLM)-based agents inherit these shortcomings through their overreliance on heuristic pattern matching, yielding recommendations prone to shallow correlation bias, limited causal inference, and brittleness in sparse-data scenarios. We introduce STARec, a slow‐thinking augmented agent framework that endows recommender systems with autonomous deliberative reasoning capabilities. Each user is modeled as an agent with parallel cognitions: fast response for immediate interactions and slow reasoning that performs chain-of-thought rationales. To cultivate intrinsic slow thinking, we develop anchored reinforcement training—a two-stage paradigm combining structured knowledge distillation from advanced reasoning models with preference-aligned reward shaping. This hybrid approach scaffolds agents in acquiring foundational capabilities~(preference summarization, rationale generation) while enabling dynamic policy adaptation through simulated feedback loops. Experiments on MovieLens 1M and Amazon CDs benchmarks demonstrate that STARec achieves substantial performance gains compared with state‐of‐the‐art baselines, despite using only 0.4\% of the full training data.
\end{abstract}

\ccsdesc[500]{Information systems~Recommender systems}

\keywords{Recommendation Agents, Deliberate Reasoning, Reinforcement Learning}

\maketitle

\section{Introduction}

Recommender systems are pivotal in modern information environments, guiding users through vast categories of items such as products, articles, or services. 
However, conventional approaches remain constrained by their reliance on historical interaction patterns and rigid feature engineering.
Although effective in narrow domains, these systems lack the cognitive flexibility to interpret open-world knowledge, infer latent preferences from natural language, or adapt to evolving user motivations, limits that become particularly acute in cold-start scenarios or complex decision-making contexts. The emergence of large language models~(LLMs) has ignited transformative potential, promising to bridge this gap through their unparalleled semantic reasoning and open-domain knowledge. Yet, as we demonstrate, unlocking truly deliberative recommendation capabilities requires moving beyond ``fast-thinking'' LLM agents toward systems capable of human-like slow reasoning.

Recent work in LLM-based recommendation agents~\cite{zhangAgentCFCollaborativeLearning2023,zhangGenerativeAgentsRecommendation2024} have demonstrated initial success in parsing user historical behaviors as user profiles and interacting with candidate items. However, these systems predominantly operate in a reactive ``System 1'' mode~\cite{kahneman2011thinking}, relying on heuristic pattern matching between user inputs and items. This manifests in three critical shortcomings: (1) shallow correlation capture rather than causal preference modeling, (2) limited capacity for multi-step inference to reconcile conflicting user signals, and (3) brittleness when handling sparse or ambiguous interaction histories. The absence of deliberate ``System 2'' reasoning~\cite{kahneman2011thinking} characterized by conscious preference decomposition, counterfactual evaluation, and iterative refinement—results in recommendations that lack personalization depth and long-term utility alignment significantly.

Typically, introducing deliberate reasoning through reinforcement learning~(RL) faces distinct technical challenges. First, the combinatorial nature of recommendation action spaces exacerbates RL’s cold-start problem, which is a challenge recently reported in DeepSeek-R1’s technical report~\cite{deepseek-aiDeepSeekR1IncentivizingReasoning2025a}. Second, conventional reward designs~(\eg CTR maximization) poorly align with the delayed, multifaceted satisfaction inherent to human decision-making. Third, the distributional shift between LLMs’ pretraining data~(general web corpora) and recommendation-specific reasoning patterns creates a semantic gap that standard RL fails to address. These challenges necessitate a fundamentally new training paradigm that scaffolds slow reasoning while maintaining sample efficiency.

To address the aforementioned limitations, in this paper, we introduce STARec, a \underline{S}low-\underline{T}hinking \underline{A}ugmented agent framework designed to endow \underline{Rec}ommender systems with autonomous deliberate reasoning capabilities. 
Specifically, our approach centers on modeling each user as an autonomous agent equipped with dual-process cognition that employs fast and intuitive thinking for personalized ranking, and slow thinking for deliberate reasoning.
These agents are engineered to dynamically learn and refine their understanding of user preferences through an autonomous cycle encompassing interaction with items, processing of behavioral feedback, and a sophisticated self-reflection mechanism. Therefore, this approach moves beyond static user profiles, enabling a continuous adaptation to evolving user tastes.

In order to cultivate the intrinsic reasoning capabilities of the agents, we propose anchored reinforcement training that bridges the gap between LLMs' generic knowledge and domain-specific slow reasoning requirements with two specific training stages. 
In the first stage, we implement structured knowledge distillation from a teacher model with slow-thinking capability~(\eg DeepSeek-R1), which instills foundational capabilities in the agents, including user preference summarization, initial item ranking logic, and the generation of chain-of-thought~(CoT) rationales.
The second stage introduces an RL paradigm to further optimize the agents' ranking decisions. We formulate a ranking-oriented reward modeling strategy to simultaneously guide both the generation of ranking lists and the updating of preference summarization during the agent pipeline.
Through iterative interaction with simulated user feedback loops, the agents learn to dynamically adjust their CoT generation and ranking policies, achieving precise alignment with evolving preference landscapes.

Our contributions are summarized as follows:
\begin{itemize}
    \item We propose STARec, a novel LLM-based agent framework where individual user agents autonomously learn and reason to model user preferences and acquire recommendation-specific knowledge through a dual-process cognitive architecture.
    \item We devise the anchored reinforcement training strategy that synergistically combines SFT anchoring with knowledge distillation from a powerful reasoning model for foundational capabilities, and RL with user preference alignment for enhancing preference-aware CoT reasoning.
    \item We demonstrate through extensive experiments on the ML-1M and Amazon CDs benchmarks that STARec significantly enhances recommendation performance that surpasses state-of-the-art baselines even though trained on only 0.4\% of the full training data amount.
\end{itemize}

\section{Related Work}

\subsection{LLM-Based Agents}

LLM-based agents are increasingly recognized as a crucial research area and a potential pathway to artificial general intelligence~\cite{chengExploringLargeLanguage2024a,xiRisePotentialLarge2023,LLMSurvey}. These agents exhibit strong generalization capabilities through natural language interfaces, making them applicable across a wide array of fields~\cite{wangSurveyLargeLanguage2024}, and their development is supported by conceptual frameworks like the ``Cognition-Planning-Feedback'' and ``Brain-Perception-Action'' models~\cite{chengExploringLargeLanguage2024a,xiRisePotentialLarge2023}. Research efforts aim to enhance the capabilities of individual agents by exploring methods such as synergizing reasoning and action~\cite{yaoReActSynergizingReasoning2023}, integrating LLMs with personal data~\cite{liPersonalLLMAgents2024}, and leveraging memory and experience for better decision-making~\cite{zhaoExpeLLLMAgents2024,parkGenerativeAgentsInteractive2023}. As research progresses, multi-agent collaboration is emerging as a key approach for handling complex tasks; initial frameworks~\cite{talebiradMultiAgentCollaborationHarnessing2023} have paved the way for numerous successful applications~\cite{chenAutoAgentsFrameworkAutomatic2024,kannanSMARTLLMSmartMultiAgent2024,qianChatDevCommunicativeAgents2024}. Advanced systems like AgentVerse~\cite{chenAgentVerseFacilitatingMultiAgent2023}, MetaGPT~\cite{hongMetaGPTMetaProgramming2024}, and AutoGen~\cite{wuAutoGenEnablingNextGen2023} further push the boundaries by exploring human-inspired group dynamics, encoding workflows, and enabling conversational collaboration, demonstrating the power of multi-agent systems in solving intricate problems.

\subsection{LLM for Recommender Systems}

The application of LLM in recommender systems represents a profound research paradigm shift, primarily following two technical routes: building direct, end-to-end LLM-based recommendation frameworks~\cite{cuiM6RecGenerativePretrained2022,gengRecommendationLanguageProcessing2023a,gengBreakingLengthBarrier2024,wangCanSmallLanguage2024,houLargeLanguageModels2024,wangRecMindLargeLanguage2024,yangItemLanguageModelConversational2024} and using LLMs to enhance traditional recommendation models~\cite{jiaLEARNKnowledgeAdaptation2024,sunLargeLanguageModels2024,xiOpenWorldRecommendationKnowledge2023,zhangGenerativeAgentsRecommendation2024,zhangAgentCFCollaborativeLearning2023}. These approaches expand the technical boundaries and offer solutions for long-standing issues like data sparsity and cold starts. End-to-end systems aim to convert recommendation tasks into language modeling problems, unifying processes~\cite{cuiM6RecGenerativePretrained2022,gengRecommendationLanguageProcessing2023a}, improving efficiency through strategies like hierarchical encoding or knowledge distillation~\cite{gengBreakingLengthBarrier2024,wangCanSmallLanguage2024}, enabling zero/few-shot recommendations with advanced prompts~\cite{houLargeLanguageModels2024,wangRecMindLargeLanguage2024}, and aligning item data with pre-trained knowledge~\cite{yangItemLanguageModelConversational2024}. Meanwhile, enhancement approaches focus on integrating LLMs' open-world knowledge by fusing it with collaborative signals~\cite{jiaLEARNKnowledgeAdaptation2024}, adding reasoning~\cite{sunLargeLanguageModels2024}, mapping preferences~\cite{xiOpenWorldRecommendationKnowledge2023}, or utilizing generative agents to build user simulation systems for training in sparse-data scenarios~\cite{zhangGenerativeAgentsRecommendation2024,zhangAgentCFCollaborativeLearning2023}.

\subsection{Deliberate Reasoning in LLM}

The reasoning capabilities of LLM are evolving from intuitive System 1 responses towards more deliberate System 2 processes~\cite{liSystem1System2025}. This transition is fundamentally supported by the CoT method~\cite{wuChainThoughtPrompting2023}, which enhances reasoning transparency and accuracy by guiding models to generate intermediate steps, thereby stimulating the use of internal knowledge and advancing AI towards more profound cognitive models~\cite{liSystem1System2025, wuChainThoughtPrompting2023}. Despite its benefits, CoT's linear structure struggles with complex problems, leading to the development of frameworks like Tree of Thoughts (ToT)~\cite{yaoTreeThoughtsDeliberate2023} and Graph of Thoughts (GoT)~\cite{bestaGraphThoughtsSolving2024}. These newer approaches break linear constraints by allowing models to explore multiple reasoning paths or arbitrary graph structures, improving performance in tasks requiring planning or synergistic thought combination. Further advancing cognitive modeling, Meta CoT~\cite{xiangSystem2Reasoning2025} focuses on cultivating metacognitive abilities by modeling the reasoning processes themselves. Concurrently, Reinforcement Learning (RL) has shown significant promise in enhancing these capabilities, with models like DeepSeek-R1~\cite{deepseek-aiDeepSeekR1IncentivizingReasoning2025a} demonstrating strong reasoning through large-scale RL training, and Kimi k1.5~\cite{teamKimiK15Scaling2025a} focusing on long-context optimization, although challenges such as multimodal integration and training stability persist and require future research.

\section{Preliminaries}

\subsection{Problem Formulation}

Recommender systems are pivotal in navigating vast information spaces by suggesting items (\eg products, articles, or services) that are likely to be of interest to users. Typically, a recommender system involves a set of users, denoted as $\mathcal{U}$, and a set of items, denoted as $\mathcal{I}$. Users engage with items through various forms of interaction, such as clicks, purchases, or ratings. These user-item interactions are fundamental, as they provide explicit or implicit signals of user preferences. The core objective of a recommender system is generally to predict a user's preference for items they have not yet interacted with, or to generate a ranked list of items, prioritizing those most likely to align with the user's interests.

Building upon this foundational understanding, the recommendation task is framed as a ranking problem. For a given user $u$, the system considers their evolving preference profile, denoted as $P_u$.This profile, $P_u$, is conceptualized as a textual description generated by a large language model~(LLM) to encapsulate the user's tastes and interests. Given this profile and a set of candidate items $C = \{ c_1, c_2, \dots, c_n \}$, the system aims to generate a ranking list $\hat{R} = \{ \hat{r}_1, \hat{r}_2, \dots, \hat{r}_n \}$. This list is designed to optimally reflect the user's underlying preferences, prioritizing items of genuine interest. In our method, the preference profile $P_u$ is not static, which is iteratively updated based on feedback from recommendation agent interactions and an internal self-reflection process, with the goal of continuously enhancing the quality of $\hat{R}$. The core challenge lies in enabling the LLM agent to accurately infer and adapt its internal representation of $P_u$, and subsequently translate this understanding into effective item rankings.

\subsection{Reinforcement Learning for LLM-based Recommendation}

Reinforcement learning~(RL) provides a principled methodology for optimizing LLM agents in sequential decision-making tasks requiring explicit deliberation. Unlike traditional supervised learning that mimics static patterns, RL enables dynamic policy adaptation through environmental interactions—a critical capability for intelligent systems balancing exploration with structured reasoning. This paradigm proves particularly effective for enhancing LLMs' slow thinking abilities through reward-driven iterative refinement. In our recommendation agent scenario, such capabilities manifest as progressive inference of user preference signals and deliberate evaluation of predicted items.

Distinct from learning paradigms that rely solely on imitating static datasets, RL facilitates the direct optimization of an agent's policy $\pi_\theta$ by learning from interactions and feedback, which can originate from a downstream system like a recommender system. This direct optimization allows the LLM's policy to continuously adapt and refine its decision-making capabilities without extensive dependence on manually labeled supervision. Such a methodology seeks to align the LLM's generation process with specific objectives, for instance, maximizing recommendation performance by improving the agent's ranking decisions for a more precise alignment with genuine user preferences. Generally, the optimization objective in this RL framework is to find a policy $\pi_\theta(a|s)$ that maximizes the expected reward as:
\begin{equation}
  \label{eq:rl}
  \mathcal{L}_{RL}(\theta) = E_{s \sim p(s), a \sim \pi_{\theta}(a|s)}[f(a|s)].
\end{equation}
This objective aims to maximize the expected value of the reward $f(a|s)$ (\eg a recommendation quality metric like NDCG) over inputs $s$ sampled from an empirical distribution $p(s)$. Outputs $a$ denotes structured outputs of ranking decisions or item recommendation lists generated by $\pi_\theta(a|s)$. 

This framework endows our recommendation agents with two crucial capabilities: (1) Deliberate refinement of candidate recommendation lists through reward-aware reasoning steps~(\eg balancing immediate accuracy with long-term user satisfaction), and (2) Strategic exploration constrained by user interest distributions and item relevance priors. The reward function $f(\cdot)$ serves as a differentiable interface between generative capabilities and recommendation objectives, translating quality metrics into optimizable signals while preserving semantic coherence in recommendations.

\section{Methodology}

\begin{figure}
  \centering
  \includegraphics[width=\linewidth]{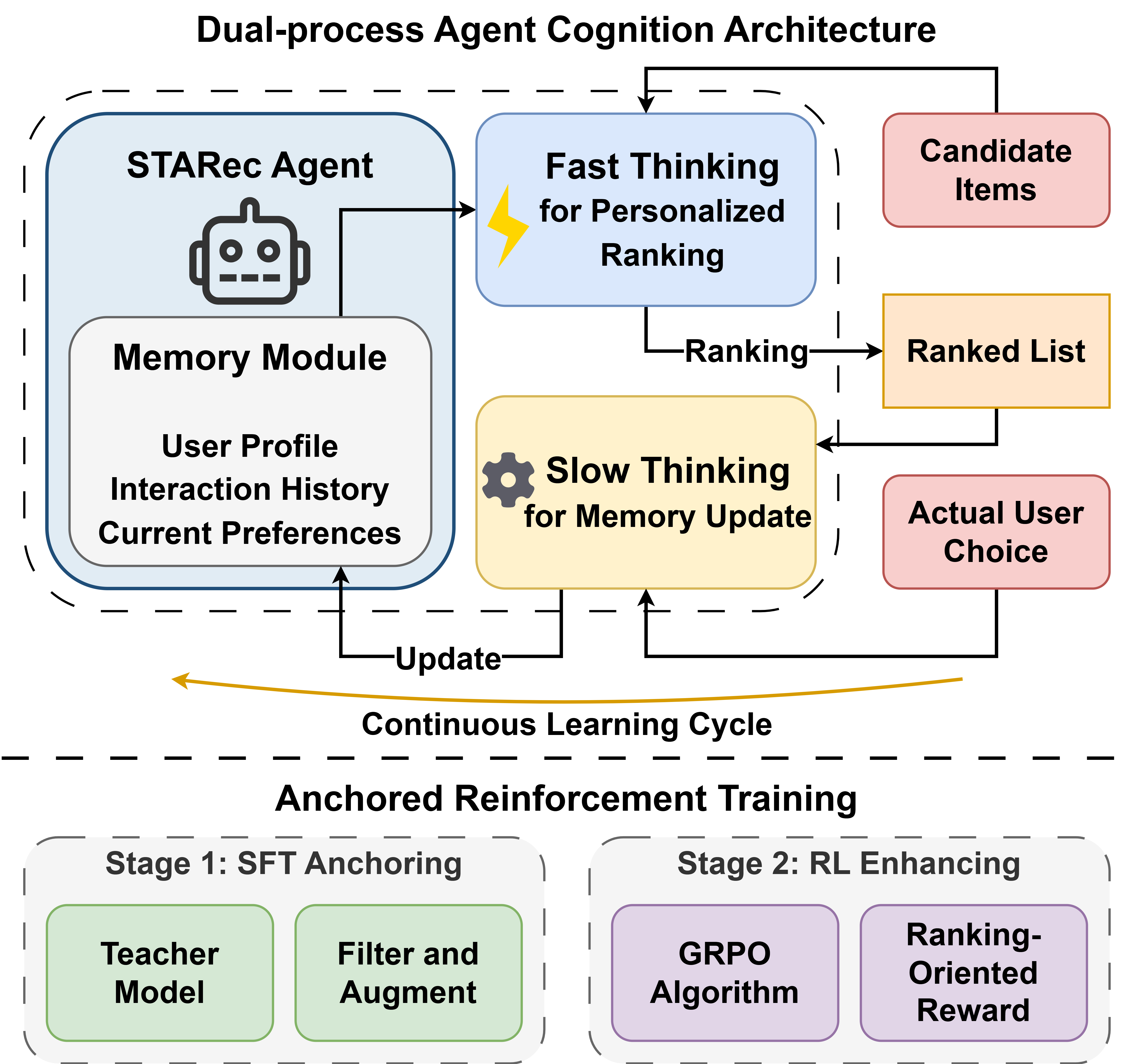}
  \caption{Overview of the proposed STARec framework.}
  \label{fig:framework}
\end{figure}

In this section, we propose STARec, a slow-thinking-augmented agent framework for recommendation systems. Our approach enables LLM-powered agents to model user preferences and acquire recommendation-specific knowledge through a dual-process cognitive architecture. Specifically, agents employ fast thinking for intuitive analysis and slow thinking for deliberate reasoning via an autonomous learning cycle. To further enhance their reasoning abilities, we introduce anchored reinforcement training, integrating training feedback from a powerful large reasoning model (via distillation-based supervised fine-tuning~(SFT)) and explicit ranking results (via reinforcement learning with ranking-based reward modeling). 
We begin by describing the dual-process agent cognition architecture, followed by a detailed explanation of the anchored reinforcement training method. The overall framework of the proposed STARec is illustrated in Figure \ref{fig:framework}.

\subsection{Dual-process Agent Cognition Architecture}
When building LLM-powered agents, it is essential to design a memory module that not only stores the initial profile but also accumulates information from the environment, while also enhancing the agent’s reasoning abilities to support goal-directed decision-making in dynamic settings. In what follows, we first introduce the memory design, followed by a discussion of the reasoning patterns.

\subsubsection{Memory Architecture}
To endow the agent with personalized insights, we equip each agent with a memory module in the form of LLM-readable natural language text. Specifically,
this module stores and manages user-specific data, including historical interactions and its current, nuanced understanding of user potential preferences. The memory can be continuously updated through ongoing interactions and reflective learning, thereby dynamically capturing evolving user tastes.

\subsubsection{Fast Thinking for Personalized Ranking}
Given the instantiated agents, we employ them to deliver personalized recommendations, focusing on the ranking task in this paper. Specifically, during this task, the agent is presented with a set of candidate items and is prompted to rank them using its memory through fast thinking. The prompt includes: (1) user demographic information (e.g., gender, age, occupation, when available, as in datasets like ML-1M); (2) a description of the user's current preferences; (3) historical interactions, including item titles, associated metadata (e.g., movie release year and genres, or product brands), and any prior feedback; and (4) the candidate items along with their attributes. The LLM processes this comprehensive input to produce a ranked list of items, accompanied by explanations that reflect the agent’s current reasoning about the user’s preferences. Table \ref{tab:interactive_ranking_io} outlines the input components provided to the LLM and the expected output format for the ranking task.

\begin{table}
  \caption{Input and output of the interactive ranking task.}
  \label{tab:interactive_ranking_io}
  \begin{tabular}{lm{0.75\linewidth}}
    \toprule
    \multicolumn{2}{c}{\textbf{Task Input}} \\
    \midrule
    System & You are a movie recommendation system. Please rank the provided list of candidate movies according to the user's profile and preferences. \\
    \midrule
    User & User Profile: \{Gender\}, \{Age\}, \{Occupation\}, \{User Description\}, \{Viewing History\}.\newline Candidate Movies: \{Title\}, \{Year\}, \{Genres\}.\\
    \midrule
    \multicolumn{2}{c}{\textbf{Task Output}} \\
    \midrule
    Assistant & <think> [Thought Process.] </think> \newline
    1. [Movie Title] - [Brief Explanation] \newline
    2. [Movie Title] - [Brief Explanation] \newline
    ... (Continue for all candidate movies) \\
  \bottomrule
\end{tabular}
\end{table}

\subsubsection{Slow Thinking for Memory Update} 
Despite the effectiveness of LLMs, subtle misalignments can still arise between an agent's simulated reasoning and a user's true preferences. To address this, we leverage slow thinking. This process prompts the agent to retrospectively analyze discrepancies between its reasoning trajectories and the user's actual behavior. Through this analysis, the agent can uncover latent user preferences and refine its memory by integrating these new insights.

\paratitle{Behavior Analysis.} When interacting with recommender systems, users typically provide implicit or explicit feedback (\eg click, comment, and like), reflecting their preferences for the candidate item. Similarly, in our simulated ranking task, the agent also generates feedback signals based on the assigned ranks of target items: items ranked highly are treated as ``Predicted Liked'', while those ranked lower are considered ``Predicted Disliked''. We then compare these simulated feedback signals with actual user behavior to identify discrepancies and assess the alignment between the agent’s predictions and real user preferences. For instance, if an item receives a high ranking from the agent but negative feedback from the user, this suggests the agent fails to accurately capture the user's real interests. In such cases, the agent’s memory should be further updated to better reflect user preferences for personalized simulation.

\paratitle{Self-Reflection.} 
To address discrepancies between user feedback and agent predictions, we prompt the agent to engage in self-reflection, using slow thinking to align its memory with the user’s true preferences.
Specifically, we construct a reflective query for the LLM that incorporates four key components: (1) the agent's current memory of user preferences; (2) details of the candidate item; (3) the agent's original prediction; and (4) the user's actual feedback.
Table \ref{tab:memory_update_io} presents the inputs to the LLM for this reflection task and the expected output. By processing this reflective query, the LLM generates an updated preference summary that integrates new information, reconciles inconsistencies, and better captures the user’s evolving interests. 

This iterative cycle of prediction, comparison, self-reflection, and memory update enables the agent to continuously learn from its errors and adapt to the user's evolving interests. Finally, the agent's interaction record is updated with the current item and its associated feedback, providing valuable context for all future summarization and ranking tasks.

\begin{table}
  \caption{Input and output of the memory update task.}
  \label{tab:memory_update_io}
  \begin{tabular}{lm{0.75\linewidth}}
    \toprule
    \multicolumn{2}{c}{\textbf{Task Input}} \\
    \midrule
    System & You are a movie preference analyst. Your task is to analyze user movie preferences and dislikes and refine their stated preference profile. You will compare the user's prediction about liking a specific movie *before* watching it with their actual feedback *after* watching it. Use any discrepancies found to update and improve the accuracy of the user's stated preference description. \\
    \midrule
    User & User Profile: \{Gender\}, \{Age\}, \{Occupation\}, \{Current User Description\}, \{Viewing History\}. \newline
    Target Movie Information: \{Title\}, \{Year\}, \{Genres\}. \newline
    \{System's Prediction\}, \{User's Actual Feedback\}. \\
    \midrule
    \multicolumn{2}{c}{\textbf{Task Output}} \\
    \midrule
    Assistant & <think> [Thought Process.] </think> \newline
    Updated User description: [...] \\
  \bottomrule
\end{tabular}
\end{table}

\subsection{
\label{sec:training_strategy}
Anchored Reinforcement Training
}
Achieving slow thinking via only zero-shot prompts is non-trivial. Therefore, to further enhance the reasoning abilities of LLM-powered agents, we introduce anchored reinforcement training with two stages: SFT anchoring and RL enhancement.

\subsubsection{SFT Anchoring}
The first step in conducting effective supervised fine-tuning (SFT) is acquiring high-quality training data. To this end, we use a strong LLM as a teacher model to generate diverse reasoning samples. For a wide range of representative user preference scenarios and their corresponding item lists, this teacher model produces optimal ranking outputs, detailed CoT rationales, and user preference descriptions. We then apply knowledge distillation by fine-tuning the agent on this curated data, which allows it to internalize the logic behind inferring user preferences and initially ranking candidate items.

To further enhance both the quality and diversity of our dataset, we implement a refined filtering process composed of two core stages: meticulous screening and targeted augmentation. The first stage, screening, aims to remove inconsistent or low-quality data. Initially, automated scripts validate format consistency by removing samples that deviate from predefined output requirements, such as missing CoT rationales, absent rankings or preference-related keywords, or lacking a positively identified item. Subsequently, to ensure informativeness and correctness, we evaluate the Normalized Discounted Cumulative Gain (NDCG) of each sample and retain only those with higher scores. The second stage, augmentation, iteratively improves data quality through a ``prompt error + rethink'' strategy. This process involves identifying flawed examples from the initial generation, such as rankings misaligned with user preferences. For these cases, we provide targeted feedback to the teacher model and prompt it to regenerate responses accordingly, creating an iterative correction loop that further refines the data. Together, these screening and augmentation steps help construct a cleaner, more diverse, and more challenging SFT dataset. This refined corpus provides a strong foundation for the STARec agent to develop a deep understanding of user preferences, engage in complex reasoning, and adhere to structured output formats. The SFT process can be optimized as follows:
\begin{equation}
  \label{eq:sft}
  \mathcal{L}_{SFT}(\Phi) = \sum_{(x,y) \in \mathcal{Z}} \sum_{t=1}^{|y|} \log \left(P_{\Phi} \left(y_t | x, y_{<t} \right) \right),
\end{equation}
where $x$ and $y$ denote the ``Task Input'' and ``Task Output'' in the instruction tuning data, respectively; $y_t$ is the $t$-th token of $y$; $y_{<t}$ represents the tokens before $y_t$; $\Phi$ is the original parameters of the model; and $\mathcal{Z}$ denotes the training set.

\subsubsection{RL Enhancement} 
In addition to employing SFT to help the model memorize fixed reasoning patterns, we further use RL to enable the model to encourage exploration of more flexible reasoning strategies. Notably, SFT provides the model with a solid foundation in task structure, preference expression, and reasoning patterns, offering a more effective starting point for RL compared to general pre-trained models or training from scratch. In this part, we first present the RL algorithm, and then describe the reward design.
 
\paratitle{RL Algorithm.}
This RL phase, building upon the SFT model, utilizes Group Relative Policy Optimization (GRPO)~\cite{shao2024deepseekmathpushinglimitsmathematical}, as adopted by DeepSeek-R1~\cite{deepseek-aiDeepSeekR1IncentivizingReasoning2025a}, for policy optimization. GRPO is selected for its ability to significantly reduce memory consumption during training compared to traditional algorithms such as Proximal Policy Optimization (PPO)~\cite{schulman2017proximalpolicyoptimizationalgorithms}, while maintaining competitive performance. A key characteristic of GRPO is its capacity to learn directly from a rule-based reward function $f(a|s)$, derived from standard evaluation metrics, thus circumventing the need for a separate, explicitly trained reward model. This approach helps mitigate reward hacking and avoids biases potentially introduced by an auxiliary reward model. Additionally, group-based normalization within GRPO effectively addresses potential reward scaling issues. Finally, to ensure training stability, the GRPO algorithm employs a Kullback-Leibler (KL) divergence penalty (term $\mathbb{D}_{KL} \left[ \pi_{\theta} \| \pi_{ref} \right]$ in Equation \ref{eq:grpo}). This penalty regularizes policy updates, preventing excessive deviation from the reference policy (\ie the SFT model) and ensuring the retention of previously learned knowledge. GRPO optimizes policies by encouraging the generation of solutions similar to successful ones and discouraging ineffective solutions, primarily by maximizing the relative advantage within groups of generated samples. This group-based learning method enhances training stability and efficiency compared to traditional pairwise comparison methods. The objective function for GRPO is defined as:
\begin{equation}
\label{eq:grpo}
\begin{split}
  \mathcal{J}_{GRPO}(\theta) =& \mathbb{E} \left[ q \sim P(Q), \{o_i\}_{i=1}^G \sim \pi_{\theta_{old}}(O|q) \right] \\
  & \frac{1}{G} \sum_{i=1}^G \frac{1}{|o_i|} \sum_{t=1}^{|o_i|} \Bigg\{ \min \left[ \frac{\pi_{\theta}(o_{i,t}|q, o_{i,<t})}{\pi_{\theta_{old}}(o_{i,t}|q, o_{i,<t})} \hat{A}_{i,t}, \right. \\
  & \left. \text{clip} \left( \frac{\pi_{\theta}(o_{i,t}|q, o_{i,<t})}{\pi_{\theta_{old}}(o_{i,t}|q, o_{i,<t})}, 1-\epsilon, 1+\epsilon \right) \hat{A}_{i,t} \right] - \\
  & \beta \mathbb{D}_{KL} \left[ \pi_{\theta} \| \pi_{ref} \right] \Bigg\}.
\end{split}
\end{equation}

\paratitle{Ranking-Oriented Reward Modeling.}
The reward function, $f(a|s)$, measures how effectively the agent performs its actions in each training iteration. In our framework, the LLM agent is responsible for two key tasks: (1) generating a ranked list of candidate items along with corresponding justifications, and (2) producing an updated preference summary in response to a reflective prompt. For the ranking task, the reward is determined by the position of the positive item in the generated list. If the positive item is ranked 1st, the agent receives a reward of +1.0. A ranking between 2nd and 5th yields a reward of +0.5, while a position between 6th and 10th results in a neutral reward of 0.0. If the item falls between 11th and 20th, a penalty of -0.5 is applied. If the positive item is not ranked within the top 20, the agent receives a penalty of -1.0. These intervals and reward values are inspired by the NDCG metric, reflecting its emphasis on ranking relevant items higher to promote effective recommendation quality. For the memory updation task, it is non-trivial to assess the effectiveness of an updated preference. Here, we introduce an indirect approach. Specifically, after the agent generates a new preference summary, it is immediately used in a follow-up item ranking task involving a relevant candidate set. The reward 
$f(a|s)$ is then computed based on the performance of this subsequent ranking, using the same reward schema as in the primary ranking task. This approach encourages the agent to generate preference summaries that enhance downstream recommendation performance. Finally, GRPO leverages the scalar reward signals derived from both tasks, to iteratively update the agent’s parameters. This training process gradually guides the agent to produce higher-quality outputs (\ie ranked lists and preference summaries), resulting in significant improvements in recommendation accuracy and personalization for interactive ranking scenarios.

\section{Experiments}
In this section, we first introduce the settings in our experiment, then present the main results together with in-depth analyses to elucidate our findings further.

\subsection{Experimental Setup}

\subsubsection{Dataset}

We conducted our experiments on two widely used public datasets, including \textit{MovieLens 1M}~(ML-1M) and \textit{Amazon CDs and Vinyl}~(CDs). For both datasets, user interactions were chronologically ordered, and interaction sequences were truncated to a maximum length of 40. Following standard practices, we filtered out users and items with fewer than 10 interactions. Subsequently, to manage computational resources, subsets comprising 1,000 users for training and 1,000 users for testing were sampled from each dataset. Interactions were defined based on user ratings, with ratings exceeding 3 classified as positive. For the ML-1M dataset, features comprised user metadata (gender, age, occupation) and item metadata (title, genre, release year). For the CDs dataset, features comprised item attributes such as title and band name. The statistics of these processed datasets are summarized in Table~\ref{tab:datasets}.

\begin{table}
  \caption{Statistics of the datasets after preprocessing.}
  \label{tab:datasets}
  \begin{tabular}{lcccc}
    \toprule
    \textbf{Dataset} & \textbf{\#Users} & \textbf{\#Items} & \textbf{\#Inters.} & \textbf{Sparsity} \\
    \midrule
    ML-1M (Full) & 6,040 & 3,883 & 1,000,209 & 95.74\% \\
    \quad -Sample & 1,000 & 2,739 & 40,000 & 98.54\% \\
    \midrule
    CDs (Full) & 1,944,316 & 544,442 & 4,543,369 & 99.99\% \\
    \quad -Sample & 1,000 & 29,483 & 40,000 & 99.86\% \\
  \bottomrule
\end{tabular}
\end{table}

\subsubsection{Evaluation Metric}

Performance is evaluated using the Normalized Discounted Cumulative Gain at K (NDCG@K) metric, with K set to 1, 5, 10, and 20. Consistent with established methodologies, we employ a leave-one-out strategy for evaluation. In this setup, the last item of each historical interaction sequence is designated as the ground-truth. The model is then tasked with ranking this ground-truth item against 19 other randomly sampled items. To minimize the impact of randomness, all test instances are executed three times, and the average of these results is reported.

\subsubsection{Baseline Methods}

To evaluate the performance of the proposed method, we conducted a comparative analysis against a comprehensive suite of baseline models. These baselines include classical approaches, established sequential recommendation algorithms, and recent LLM-based methodologies.

\begin{itemize}
    \item \textbf{Pop} recommends items based on their overall popularity, ranking items with more interactions higher.
    \item \textbf{BPR} utilizes matrix factorization to learn user and item representations by optimizing the BPR loss function.
    \item \textbf{GRU4Rec} utilizes Gated Recurrent Units to model user session sequences and predict the next item a user is likely to interact with based on their recent browse history.
    \item \textbf{SASRec} captures sequential patterns in user interaction histories using a Transformer self-attention mechanism.
    \item \textbf{LLMRank} utilizes LLMs for zero-shot ranking in recommender systems by framing recommendation as a conditional ranking task solved via prompting.
    \item \textbf{AgentCF} simulates user-item interactions by modeling users and items as collaborative agents.
\end{itemize}

The personalized recommendation capabilities of our user agents are developed and refined through training on a designated sampling dataset. To ensure a fair comparison of their performance, its performance was compared against BPR, and SASRec, which were also trained on this same sampling dataset (referred to as $\mathrm{BPR}_\mathrm{sample}$, and $\mathrm{SASRec}_\mathrm{sample}$, respectively). For comprehensive benchmarking, the performance of these baseline models trained on the complete dataset is also reported (denoted $\mathrm{BPR}_\mathrm{full}$, and $\mathrm{SASRec}_\mathrm{full}$). The Pop model, which utilizes statistical data for recommendations, undergoes training on the complete dataset. Conversely, as the LLMRank and AgentCF methods do not necessitate a training phase, their performance is directly evaluated on the sampled dataset.

\subsubsection{Implementation Details}

We selected DeepSeek-R1-Distill-Qwen-32B as the teacher model for generating our SFT dataset. The base models for our user agents were primarily Qwen2.5-7B, with Qwen2.5-0.5B and Qwen2.5-1.5B used in some analysis experiments. For the SFT phase, models were trained using the Llama-Factory framework~\cite{zheng2024llamafactory}. We employed a learning rate of 1.0e-4 for the 0.5B and 1.5B models, and 1.0e-5 for the 7B model. The maximum sequence length for SFT inputs was 16384 tokens. The SFT process spanned 3 epochs for all model sizes. The SFT data was generated by prompting the teacher model. The RL phase employed the GRPO algorithm, implemented using the VeRL framework~\cite{sheng2024hybridflow}. During RL, a uniform learning rate of 1.0e-6 was applied to all models, training for 1 epoch. The GRPO algorithm utilized a batch size of 64, a KL divergence penalty coefficient of 1.0e-3, and a number of rollouts of 8. The maximum input/output token lengths were 4096 and 16384 respectively. For traditional baseline methods, we utilized implementations from a popular open-source recommendation framework RecBole~\cite{recbole[1.0],recbole[2.0]}.

\subsection{Performance Comparison}

\begin{table*}[t]
  \centering
  \caption{Performance comparison of various recommendation methods. Results are reported using NDCG@K for K=1, 5, 10, and 20. ``sample'' and ``full'' denote training with sampled or the full training set. The best performance score is denoted in bold, with the second best underlined.}
  \label{tab:model_performance_wide_centered}
  \begin{tabular}{lcccccccc}
    \toprule
    \multirow{2}{*}{\textbf{Method}} & \multicolumn{4}{c}{\textbf{ML-1M}} & \multicolumn{4}{c}{\textbf{Amazon CDs}} \\
    \cmidrule(lr){2-5} \cmidrule(lr){6-9}
    & NDCG@1 & NDCG@5 & NDCG@10 & NDCG@20 & NDCG@1 & NDCG@5 & NDCG@10 & NDCG@20 \\
    \midrule
    Pop                                 & 20.50 & 41.89 & 49.08 & 52.86 & 24.10 & 43.18 & 47.69 & 55.13 \\
    $\mathrm{BPR}_\mathrm{sample}$      & 24.50 & 42.52 & 50.74 & 54.64 & 26.70 & 44.99 & 50.55 & 57.76 \\
    $\mathrm{BPR}_\mathrm{full}$        & 33.50 & 54.92 & 60.85 & 62.60 & 46.20 & 60.35 & 63.66 & 67.72 \\
    $\mathrm{GRU4Rec}_\mathrm{sample}$  & 38.00 & 60.04 & 64.65 & 66.28 & 33.80 & 46.21 & 49.82 & 57.47 \\
    $\mathrm{GRU4Rec}_\mathrm{full}$    & 54.70 & 72.97 & 75.47 & 76.46 & 60.00 & 68.43 & 70.84 & 75.22 \\
    $\mathrm{SASRec}_\mathrm{sample}$   & 42.30 & 62.59 & 66.68 & 68.53 & 34.90 & 46.40 & 49.76 & 57.93 \\
    $\mathrm{SASRec}_\mathrm{full}$     & \textbf{57.50} & \underline{73.29} & \underline{76.51} & \underline{77.47} & 62.30 & 77.13 & 79.47 & 80.32 \\
    \midrule
    LLMRank                     & 31.90 & 52.70 & 56.19 & 60.76 & 58.70 & 72.35 & 73.54 & 74.31 \\
    AgentCF                     & 40.30 & 59.88 & 65.26 & 70.28 & 64.00 & 76.19 & 79.15 & 80.37 \\
    \midrule
    STARec-Teacher (R1-32B)   & 39.00 & 57.27 & 62.35 & 65.12 & 60.00 & 74.91 & 77.18 & 78.46 \\
    STARec-SFT$_\mathrm{sample}$                       & 35.60 & 55.74 & 60.87 & 63.24 & 57.30 & 72.88 & 74.90 & 76.58 \\
    STARec$_\mathrm{sample}$                         & \underline{55.40} & \textbf{75.27} & \textbf{77.16} & \textbf{77.96} & \textbf{68.30} & \textbf{81.40} & \textbf{82.63} & \textbf{84.36} \\
    \bottomrule
  \end{tabular}
\end{table*}

The performance of various methods on the ML-1M and Amazon CDs datasets is presented in Table \ref{tab:model_performance_wide_centered}.  From these results, we draw several key insights into the effectiveness of our approach:

(1) The utilization of our dual-process agent cognition architecture integrated with the DeepSeek-R1-32B teacher model establishes a strong performance baseline, demonstrating considerable capabilities. While some traditional methods trained on sampled data, such as $\mathrm{SASRec}_\mathrm{sample}$, exhibit competitive results, the DeepSeek-R1-32B model provides a robust foundation of generalist knowledge that is crucial for the subsequent distillation and fine-tuning stages within our framework.

(2) SFT on data distilled from the teacher model proves highly effective, enabling more compact student models to successfully inherit a significant portion of the teacher's capabilities.  
Most importantly, the subsequent application of RL elevates our student models to a significantly higher level of performance. Beyond markedly surpassing their SFT counterparts and the original teacher model, our RL-enhanced models demonstrate clear advantages when compared against traditional recommendation systems. Specifically, when trained with a similar volume of data (\ie sampled datasets), our RL models consistently show superior performance over these traditional methods. 

(3) Furthermore, our models achieve a level of effectiveness that is comparable, and in certain scenarios even superior, to traditional recommenders trained on the entire full datasets. This achievement is particularly notable as our models are trained using only approximately \textit{\textbf{0.4\%} of the complete data volume}. Such strong performance, attained with remarkable data efficiency, underscores the robust generalization capabilities fostered by our RL-driven methodology.

\subsection{Further Analysis}

\subsubsection{\label{sec:ablation} Ablation Study}

To evaluate the individual contributions of key components within STARec—namely, the choice of RL algorithm, the SFT phase, and the self-reflection mechanism—ablation studies were conducted. We utilized the 1.5B model variant, selected for its effective balance between performance and computational efficiency. Experiments were performed on the ML-1M and Amazon CDs datasets, with the results detailed in Table \ref{tab:ablation}.

(1) GRPO vs. Reinforce++: The choice of RL algorithm is a critical aspect of STARec. We compared the performance of STARec using its default GRPO algorithm against a variant that employed Reinforce++. Both GRPO and Reinforce++ are considered representative RL algorithms that operate without a critic model. The results, detailed in Table \ref{tab:ablation}, indicate that both algorithms achieve largely comparable performance on the ML-1M and Amazon CDs test datasets. Our findings on these recommendation datasets suggest that both GRPO and Reinforce++ are similarly effective in terms of final recommendation accuracy. This suggests that the STARec framework demonstrates robustness to the specific selection between these two RL algorithms, both proving to be viable choices for optimizing recommendation performance.

(2) w/o SFT Anchoring: To assess the importance of SFT as an initialization phase, the base model was trained directly using RL, bypassing the SFT stage. The resultant Direct RL variant exhibited a marked decrease in performance when compared to the full model. This highlights the significance of SFT in establishing a robust foundational model for the subsequent RL phase.

(3) w/o Self-Reflection: For this variant, the LLM-driven self-reflection mechanism was substituted with the direct appending of interaction history, a modification designed to evaluate the efficacy of the advanced reflection mechanism. The removal of this sophisticated self-reflection component also led to a clear degradation in performance. This confirms the crucial nature of the agent's capacity to dynamically learn and refine its understanding of preferences through LLM-driven reflection.

\begin{table}
\caption{Results of ablation study on two datasets with parameter scale of 1.5B.}
\label{tab:ablation}
\begin{tabular}{lcccc}
    \toprule
    \multirow{2}{*}{\textbf{Method}} & \multicolumn{2}{c}{\textbf{ML-1M}} & \multicolumn{2}{c}{\textbf{Amazon CDs}} \\
    \cmidrule(lr){2-3} \cmidrule(lr){4-5}
    & N@1 & N@10 & N@1 & N@10 \\
    \midrule
    STARec          & 51.30 & 75.19 & 66.10 & 80.66 \\
    \midrule
    Reinforce++             & 49.50 & 73.87 & 65.20 & 82.54 \\
    w/o SFT Anchoring       & 26.10 & 53.97 & 41.20 & 63.57 \\
    w/o Self-Reflection     & 46.60 & 71.31 & 62.40 & 76.88 \\
    \bottomrule
\end{tabular}
\end{table}

\begin{table}[t]
  \centering
  \caption{Results on different model scales of STARec.}
  \label{tab:scaling}
  \begin{tabular}{lcccc}
    \toprule
    \multirow{2.5}{*}{\textbf{Method}} & \multicolumn{2}{c}{\textbf{ML-1M}} & \multicolumn{2}{c}{\textbf{Amazon CDs}} \\
    
    \cmidrule(lr){2-3} \cmidrule(lr){4-5} 
    & N@1 & N@10 & N@1 & N@10  \\
    \midrule
    STARec-SFT-0.5B                     & 34.70 & 59.23 &  55.10 & 71.82  \\
    STARec-SFT-1.5B                     & 33.80  & 60.77  & 56.70  & 73.73  \\
    STARec-SFT-7B                      & \textbf{35.60}  & \textbf{60.87}  & \textbf{57.30}  & \textbf{74.90}  \\
    \midrule
    STARec-RL-0.5B       & 48.80  & 70.91  & 61.80  & 77.54  \\
    STARec-RL-1.5B       & 51.30  & 75.19  & {66.10}  & {80.66}  \\
    STARec-RL-7B         & \textbf{55.40}  & \textbf{77.16} &  \textbf{68.30}  & \textbf{82.63}  \\
    \bottomrule
  \end{tabular}
\end{table}

\subsubsection{Scaling Law of STARec}
To systematically analyze the performance characteristics of STARec across varying model capacities, we conduct experiments using 0.5B, 1.5B, and 7B parameter versions, evaluating both the SFT and RL training stages. The results as summarized in Table~\ref{tab:scaling}.
First, STARec demonstrates a clear and consistent trend of monotonic performance improvement with increasing model size across both training stages. This aligns with the broader scaling laws observed in large language models~\cite{kaplan2020scaling}, reinforcing the effectiveness of our architecture in leveraging additional model capacity for better reasoning and recommendation quality.
Second, the method demonstrates remarkable parameter efficiency. Despite having 14× fewer parameters, the 0.5B model retains approximately 97\% and 88\% of the 7B model's final performance in the SFT and RL stages.
These findings underscore STARec's flexibility: it can scale to large models for high-performance applications, while remaining viable in edge or latency-sensitive deployments where inference speed and resource consumption are critical bottlenecks.

\subsubsection{Performance Analysis by User Activity Level}
\label{sec:performance_user_activity}

\begin{figure}
  \centering
  \begin{subfigure}[t]{0.48\columnwidth}
    \centering
    \includegraphics[width=\linewidth]{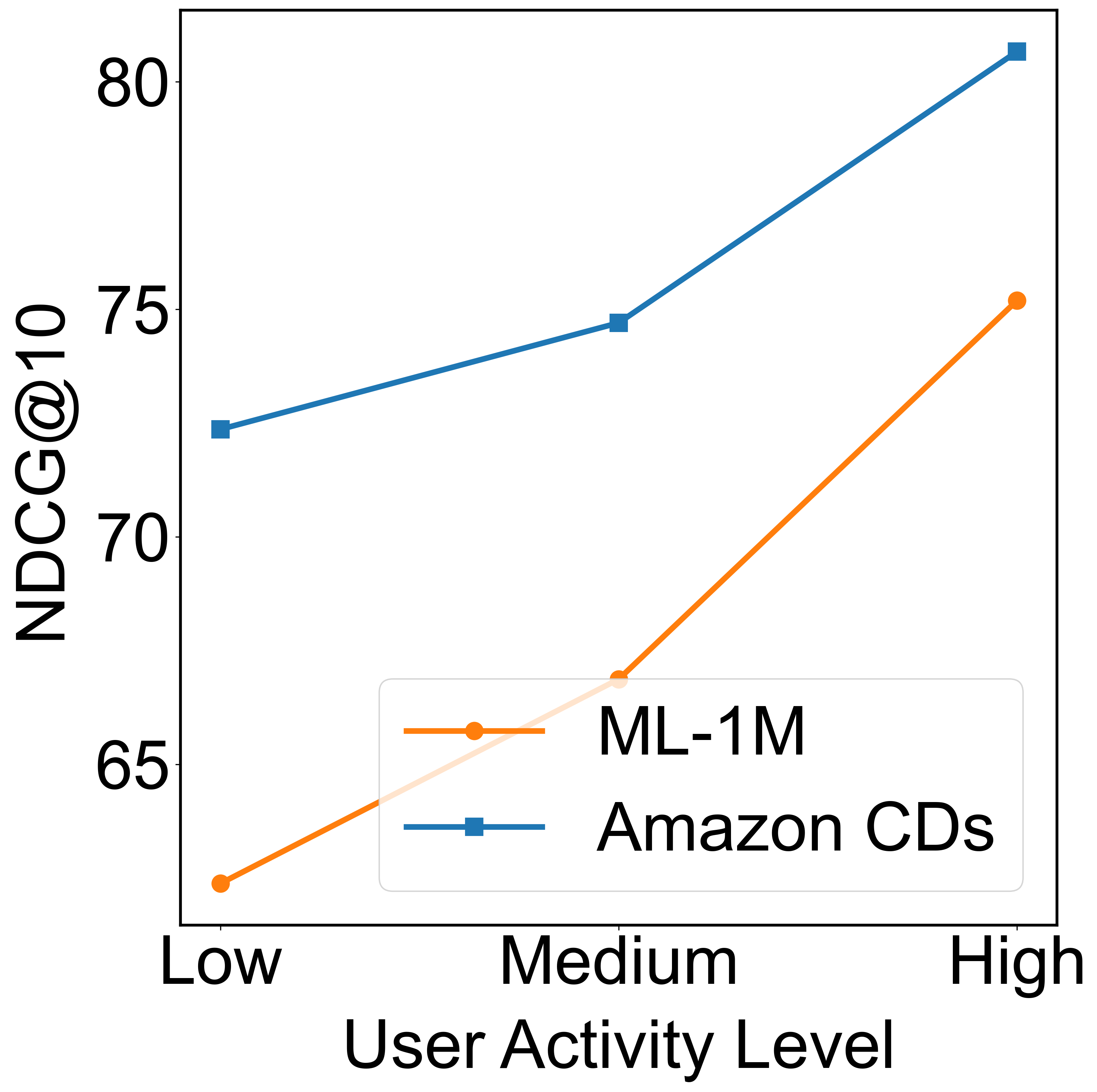}
    \caption{Results on different user activity groups.}
    \label{fig:activity-group}
  \end{subfigure}
  \hfill
  \begin{subfigure}[t]{0.48\columnwidth}
    \centering
    \includegraphics[width=\linewidth]{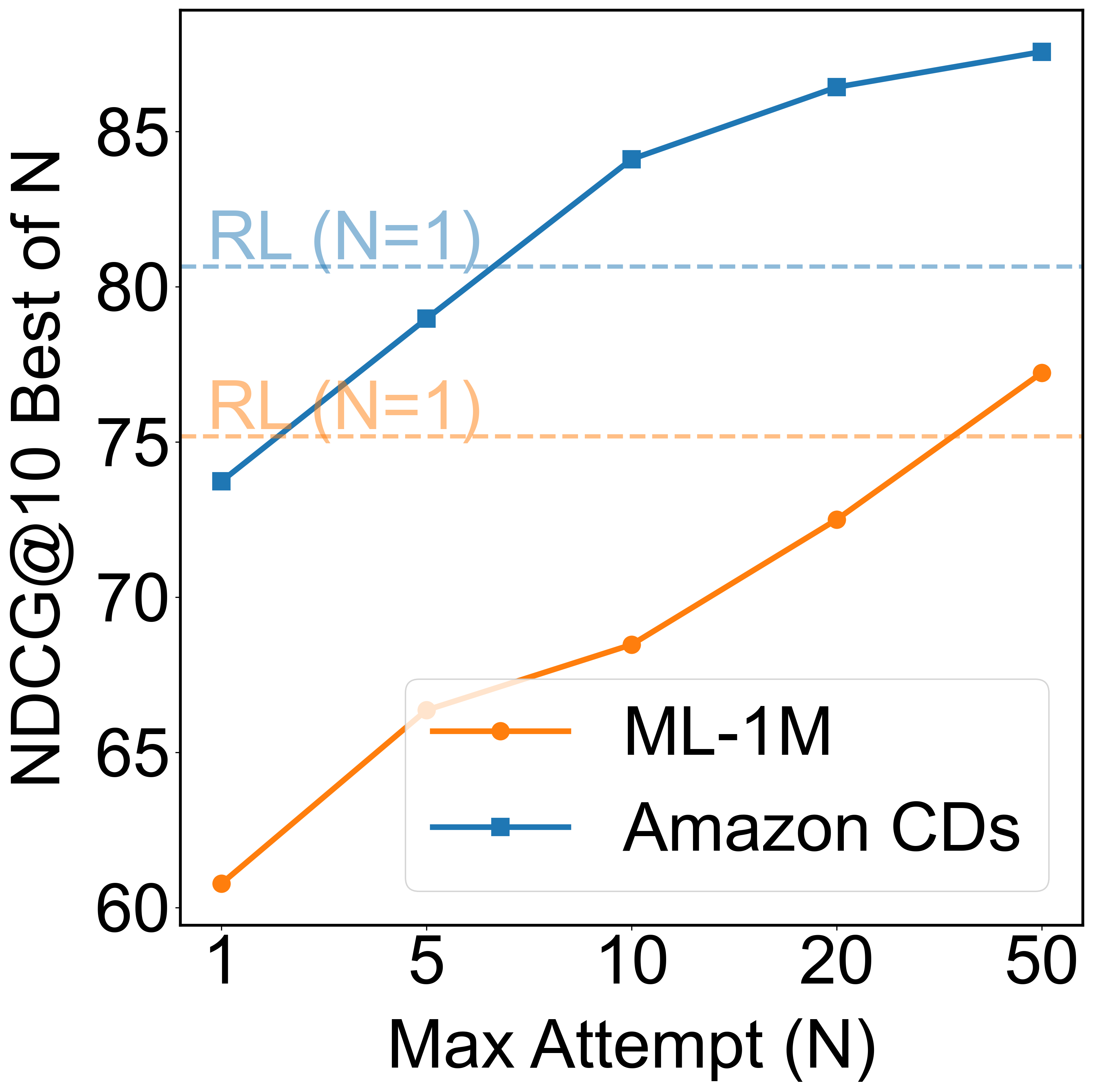}
    \caption{Best of $N$ results of SFT comparing to RL with $N = 1$.}
    \label{fig:sft_best_of_n}
  \end{subfigure}
  \caption{Performance comparisons of STARec-1.5B \wrt different settings on the ML-1M and Amazon CDs datasets.}
  \label{fig:performance-comparison}
  \vspace{-10pt}
\end{figure}

To further assess the generalization capability of the proposed STARec framework, its performance was analyzed across distinct user activity groups. The STARec-1.5B model was evaluated on these user groups for both ML-1M and Amazon CDs datasets. Users were categorized based on their historical interaction counts. Owing to differing data distributions, user activity thresholds were defined as follows.
For the ML-1M dataset, users were divided into three groups: Low Activity (10-24 interactions), Medium Activity (25-39 interactions), and High Activity (40 interactions).
For the Amazon CDs dataset, the groups were: Low Activity (10-19 interactions), Medium Activity (20-39 interactions), and High Activity (40 interactions).
Figure \ref{fig:activity-group} presents the NDCG@10 performance of the STARec-1.5B (RL) model across these activity groups.

As depicted in Figure \ref{fig:activity-group}, the performance of STARec exhibits two key characteristics: its scalability with data density and, more critically, its resilience in data-sparse environments. First, as expected, the model's performance improves with increased user activity. Users with more historical interactions (High Activity groups) achieve higher NDCG scores, as more data allows the agent to construct a more accurate user profile. However, a more significant finding is STARec's robust performance even for users with limited interaction data (Low Activity groups). Although a performance gap exists compared to more active users, the model still achieves commendable recommendation accuracy. This result strongly suggests that the framework's anchored reinforcement training strategy equips agents with effective generalization capabilities, enabling them to extrapolate user preferences and generate coherent, relevant recommendations from sparse historical data. This specific capacity to effectively address the prevalent ``cold-start'' problem for new or infrequent users is a critical differentiator, directly underscoring STARec's practical applicability.

\subsubsection{Contribution Analysis of SFT and RL}

In the STARec framework, SFT and RL fulfill distinct yet complementary functions. SFT focuses on establishing foundational capabilities, primarily through knowledge acquisition during training, whereas RL refines these capabilities using rule-based rewards. To evaluate the inherent potential of SFT models and clarify the specific contributions of RL, we introduce the ``NDCG@10 Best of N'' metric. Within the recommendation context, the NDCG@10 Best of N for an SFT model is calculated as follows: For a given user and their associated test items, we generate $n$ independent recommendation lists by sampling from the SFT model. The highest NDCG@10 score achieved across these $n$ attempts represents the user's final performance. This SFT NDCG@10 Best of N is then benchmarked against the standard NDCG@10 (N=1) performance of the corresponding RL model. We employ the STARec-SFT-1.5B model for the best of n analysis, comparing it with the STARec-1.5B model. Experiments are performed on the ML-1M and Amazon CDs datasets. For the SFT model, we assess NDCG@10 Best of N for $n = \{1, 5, 10, 20, 50\}$ using a sampling temperature of 1.0. The RL model's performance is evaluated using its standard NDCG@10 score (N=1) at a temperature of 0.2.

The results shown in Figure \ref{fig:sft_best_of_n} demonstrate the relationship between the number of sampling attempts ($N$) and the SFT model's performance relative to the RL model's standard NDCG@10. As depicted, while the SFT model underperforms the RL model when evaluated with a single sample ($N=1$), its “Best of $N$” NDCG@10 score improves substantially as $N$ increases. Notably, the SFT model's performance can approach or potentially match the RL model's standard (N=1) level given multiple opportunities. This suggests that, although the SFT model may not consistently generate the optimal recommendation in a single attempt, it often possesses the capacity to generate high-quality solutions within its potential output space. The primary challenge is reliably extracting them with a few sample attempts. Additionally, these findings indicate that the primary function of RL in the STARec framework is not to introduce entirely new knowledge but rather to perform ``success amplification.'' That is, RL acts as an optimization layer, utilizing reward signals to guide the policy towards high-reward regions pre-established by SFT. This process effectively ``sharpens'' the model's ability to select high-quality solutions, increasing the probability of generating an optimal recommendation in a single attempt (N=1). 

This perspective highlights the critical importance of our anchored reinforcement training approach in achieving high-quality recommendations.  SFT provides the essential foundation by equipping the model with fundamental capabilities, while RL delivers the targeted optimization required to consistently leverage these capabilities. This synergistic interaction is crucial for achieving effective recommendations. Furthermore, these observations hold broader relevance for LLM training, especially in specialized recommendation fields, emphasizing the necessity of a robust SFT foundation before applying RL-based optimization.

\section{Conclusion}

In this paper, we present STARec, a slow-thinking augmented agent framework that models users as autonomous LLM agents with deliberative reasoning capabilities through an autonomous learning cycle.
Our core technical contributions are twofold: 1) A novel agent architecture integrating dual-process cognition, which enables agents to conduct immediate interactions with fast thinking, while also performing slow, deliberative reasoning via self-reflection and memory updates; 2) An anchored reinforcement training paradigm that combines structured knowledge distillation to estabilish foundational reasoning capabilities with GRPO-based reinforcement learning for adaptive policy adaptation.
Extensive experiments demonstrate significant improvements on two public benchmarks, validating the framework's superiority in dynamic recommendation scenarios. It is worth noting that the generated chain-of-thought rationales further provide interpretable support for recommendation decisions, and the framework achieves strong performance even with smaller, efficiently trained models. In future work, we aim to strengthen STARec’s reasoning capabilities by integrating more advanced teacher models and adopting efficient training paradigms such as curriculum learning. We also plan to explore more sophisticated agent architectures and interactive protocols, including multi-agent collaboration, hierarchical planning, and dynamic user feedback loops, to better adapt to evolving recommendation scenarios.

\section*{GenAI Usage Disclosure}

GenAI tools were used to improve spelling, grammar, and clarity.

\begin{acks}
This work was partially supported by National Natural Science Foundation of China under Grant No. 92470205 and 62222215, Beijing Natural Science Foundation under Grant No. L233008, and Beijing Municipal Science and Technology Project under Grant No. Z231100010323009. Xin Zhao is the corresponding author.
\end{acks}

\bibliographystyle{ACM-Reference-Format}
\balance
\bibliography{main}

\end{document}